# A general mechanism of humour

**Javier Martínez**
UNED
jvermartinez@gmail.com

Abstract

*This article proposes a cognitive mechanism of humour of general applicability, not restricted to verbal communication. It is indebted to Raskin's concept of script overlap, and conforms to the incongruity-resolution theoretical framework, but it is built on the notion of constraint, an abstract correspondence between sets of data. Under this view, script overlap is an outcome of a more abstractly described phenomenon, constraint overlap. The important concept of the overlooked argument is introduced to characterise the two overlapping constraints—overt and covert. Their inputs and outputs are not directly encoded in utterances, but implicated by them, and their overlap results in another overlap at the level of the communicated utterances, that the incongruity reveals. Our hypothesis assumes as a given that the evocation of such constraints is a cognitive effect of the inferential process by which a hearer interprets utterances. We base this assumption on Hofstadter's theory of analogy-making as the essence of human thought. By substituting "stimuli" of any kind for "utterances" in this model, we obtain a mechanism as easily applicable to non-verbal communication—slapstick, cartoons— and we propose it describes the necessary and sufficient conditions for a communicative act in any modality to carry humour.*

## 1 Introduction

We know that folk theories of humour are fallacious because they describe a mysterious phenomenon in terms of other equally mysterious phenomena. What makes jokes humorous? The occurrence of something comic. But "comic" is just another way of referring to the very phenomenon we are trying to explain. A proper theory should take us out of the circle, describing humour in terms of simpler, "humourless" constituents, ultimately—and in principle—allowing a machine to detect the property "humorous" through the formal manipulation of symbols, without experiencing any subjective feeling of "humour" or "comedy".

The canned joke, conveniently brief and self-contained, has become the undisputed testing ground for humour theories, and the incongruity-resolution framework has been the prevalent one since the formulation of the Semantic Script Theory of Humour (SSTH) by Raskin (1985) and the General Theory of Verbal Humour (GTVH) by Attardo and Raskin (1991). These are, however, linguistic theories, well-suited to the analysis of canned jokes, whereas humour is a phenomenon ubiquitous in non-verbal communication. Thus, we will propose a similar mechanism resorting to Hofstadter's concept of *slippage* (Hofstadter and Mitchell 1995). This mechanism, we argue, is compatible with the SSTH/GTVH, script overlap being a necessary outcome of a more abstractly defined overlap. Such conclusion suggests the same degree of compatibility with the theory of *frame-shifting* (Coulson 2001) and related views from Cognitive Linguistics (Brône and Feyaerts 2003), and indeed with the general tenets of incongruity theories (Suls 1972; Oring 2003; Ritchie 2004: 59-64; Dynel 2009; Yus 2016) formulated since Koestler (1964), being in fact its more formal equivalent. While logically simple, it is flexible enough to serve as a general template for a resolution mechanism separating humorous from non-humorous communication in any modality, and lends itself more easily than other models to visual and non-verbal humour, as we strive to show, in the hope that our conclusions may be useful to other researchers in the field.

## 2 Methodology

Our hypothesis is concerned with humour *competence* (Raskin 1985), that is, with making explicit the "grammar" that allows a speaker to classify a stimulus as humorous. The funniness, opportunity, appropriateness, trendiness, or any other property that can be graded on a continuum from less to more, are *not* our object of analysis. One may find a joke dull and still *get it*. It is the nature of this *getting it* that we want to describe, or in Ritchie's (2018) terms, the cognitive aspect of humour *recognition*, not the psychological and socio-cultural aspects of its *appreciation*, such as enjoying it more or less, finding it more or less offensive, or deeming it more or less worth a laugh. This approach is intended to provide us with a falsifiable theory predicting what communicative acts do and do not belong to the set of well-formed humorous acts.

A second point that needs to be made is that our object of study—what makes a communicative act humorous—requires a high level of idealisation. We will abstract away from factors that would be significant if our object of study was a different one. For instance, the following two communicative acts will become virtually indistinguishable:

(1) How many Poles does it take to screw in a light bulb? Five—one to hold the light bulb and four to turn the table he's standing on. (Quoted in Attardo and Raskin 1991: 295)

(2) A scene in a musical comedy à la *Singin' in the Rain* (1952). Gene Kelly, Donald O'Connor, and Debbie Reynolds have just moved into a new apartment. They paper the walls, hang new curtains, unpack their belongings, and so on, in the most varied and ingenious ways, singing and dancing all along. At one point one of them jumps on a table and holds a lightbulb way up high. The other two grab the table by the edge from opposite sides and turn it, until the light bulb is screwed in in the ceiling. (Constructed example)

Needless to say, there are huge differences between the two, apart from the glaring one that (1) is a canned joke and (2) is a visual gag. Even in their verbalised rendition, they differ greatly in Language, Narrative Strategy, and Target (from the GTVH). But our model is not taxonomically oriented and will be blind to these differences. Like the SSTH, it will only account for the highest level of abstraction in the representation of a joke—Script Opposition and Logical Mechanism, subsuming both under a more abstractly defined cognitive phenomenon.

One other difference deserves special mention. The incongruity in (1)—the unexpected, inefficient lightbulb-rotating method—comes last, being the punchline of the joke, whereas in (2) it comes first. Our model will also be blind to the linear organisation of the text of a joke. Thus, the following two texts will get an identical description:

(3) What creature walks on four legs in the morning, two legs at noon, and three in the evening? Answer: man. (Riddle of the Sphinx)

(4) What is man? The only creature that walks on four legs in the morning, two legs at noon, and three in the evening. (Constructed example)

As can be seen, not only canned jokes, but riddles (3) and witticisms (4) will be treated as humorous texts. Section 8 will comment further on this inclusion and its significance.

We will follow a partly deductive reasoning, starting with a set of assumptions, from which we will enunciate some formal definitions, to finally postulate a general mechanism of humour and examine how this model applies in practice to a diverse group of humorous texts and non-verbal communicative acts. This being a work of theoretical elucidation—or, more humbly, "armchair theorising"—the corpus from which we draw has been collected from the scientific literature on humour we have consulted, with a few naturally occurring instances added by us.

## 3 Assumptions

### 3.1 Knowledge and communication

Our first three premises are rather uncontroversial assumptions about human communication. First, we assume that our knowledge of the world, stored in long-term memory, and our construction of the meaning of a discourse situation, in working memory, can be represented as a *semantic network*, a graph consisting of nodes—which represent concepts—connected by links—which represent semantic relations between concepts.

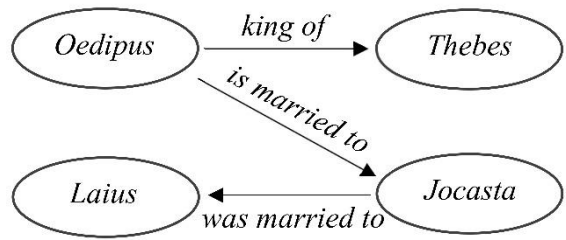

Figure 1. Semantic network

Figure 1 represents part of our knowledge of the myth of Oedipus. Nodes and links have labels in English for clarity, but they are normally not words. Hereinafter, labels in *italics* will identify concepts, and labels in "quotation marks" will identify words or other signifier units. Thus, the node *Oedipus*, representing the mythical character, is connected to the node "Oedipus" through the link *English name is*. Most nodes, though, have no link to a lexical tag, such as a melody, a particular voice quality, a face, or a nervous tic. Hence not only lexicalized concepts, but anything our memory can point to can be represented as a node.

The second assumption is that, in a discourse situation, the hearer retrieves from long-term memory a subset of her world-knowledge, and updates it with the *cognitive effects* triggered by her interpretation of the discourse. Thus, someone listening to the story of Oedipus for the first time would add to Figure 1 the links and nodes to the effect that Oedipus killed Laius, and has a daughter called Antigone, when that knowledge was communicated.

The third assumption is the elementary fact that the cognitive effects communicated by utterances are not directly encoded in those utterances, but the hearer must obtain them through inference—communication is an inferential process guided by pragmatic regularities.

## 3.2 Analogy-making and human thought

The fourth assumption is considered by his best-known proponent, Hofstadter, an "unconventional viewpoint about what thought itself is" (Hofstadter and Sander 2013: 29), so a brief explication is called for. We will not need to assume Hofstadter's strong thesis that analogy-making is the essence of human thought, but the much weaker one that analogies are constantly being implicated in human communication, and are therefore among the cognitive effects of a discourse situation.

Hofstadter's prototypical example is the following puzzle:

(5) Suppose the letter-string "abc" were changed to "abd"; how would you change the letter-string "ijk" in "the same way"? (Hofstadter and Mitchell 1995: 206)

Most people immediately notice the same abstract rule constraining the transformation of "abc" into "abd", namely, *replace the third letter with its successor in the alphabet*. The text activates in the hearer not only the pairing of two individual data, "abc" and "abd", but also a generalisable procedure to obtain one from the other, eventually allowing her to predict other pairs by analogy. The typical hearer answers that "ijk" would become "ijl" if it were changed "in the same way", even if what that "same way" *is* is not made explicit in the text, but inferred by the hearer as she interprets its intended cognitive effects. Character-string puzzles are useful because their abstract nature makes it easy to describe the process in a mathematically precise way. But analogy-making, according to Hofstadter, can be found in all domains of human activity, including the processing of the most mundane conversations:

(6) Shelley: I'm going to pay for my beer now.
Tim: Me too. (Hofstadter 1995: 76)

What does Tim mean? That he is going to pay for Shelley's beer too? Of course not. The natural interpretation of Tim's "doing the same as Shelley" is that he is going to pay for *his* beer—and it may not even be a beer, but a bottle of water. Let *f* represent that which Tim says he is going to do and happens to be the same as Shelley. *f*, so implicated, stands for *paying for Shelley's drink* when Shelley is the agent of the action, but it stands for *paying for Tim's drink* when Tim is. Thus, *f* is actually a function of an input variable, the agent, denoted *a*, and whose output is $f(a)$ = *paying for a's drink*. The pairing of *Tim* and *paying for Tim's drink* is analogous to the pairing of *Shelley* and *paying for Shelley's drink*. In this light, it is clear that Tim and Shelley are indeed going to do the "same thing", namely *f*, even if $f(Tim) \neq f(Shelley)$.

It seems safe to assume that this kind of analogy-making is ubiquitous in human communication and thought. As Hofstadter puts it, it is what makes it possible for people who hear a story to say, "Exactly the same thing happened to me!" when it is by definition impossible for the same event to happen to two different people.

But the very notion of semantic network—as in Figure 1—is proof enough that analogies are at the basis of communication and thought, for we use the same label to identify many individual links, such as the label *fatherof* on the links *Oedipus ⇨ Antigone*, *Oedipus ⇨ Ismene*, and *Laius ⇨ Oedipus* (Figure 2), making the three data pairs *analogous* to each other, their constitutive elements mapping onto one another across all pairs. Simple semantic relations are very simple analogies.

The link *Oedipus ⇨ Antigone* labelled as the semantic relation *fatherof* is expressed as *Antigone* ∈ *fatherof*(*Oedipus*), *fatherof* being the function that outputs, for a given input, the set of entities fathered by it. We use the symbol ∈ instead of = because Oedipus had children other than Antigone. Hereinafter, for simplicity and consistency of notation, outputs of functions will always be sets of nodes, even for semantic relations that point to one single node by definition— such as the relation *mothered by*—in which case the output set has only one member. It will be necessary to keep this in mind when semantic relations begin compounding, which will be the topic of the next section.

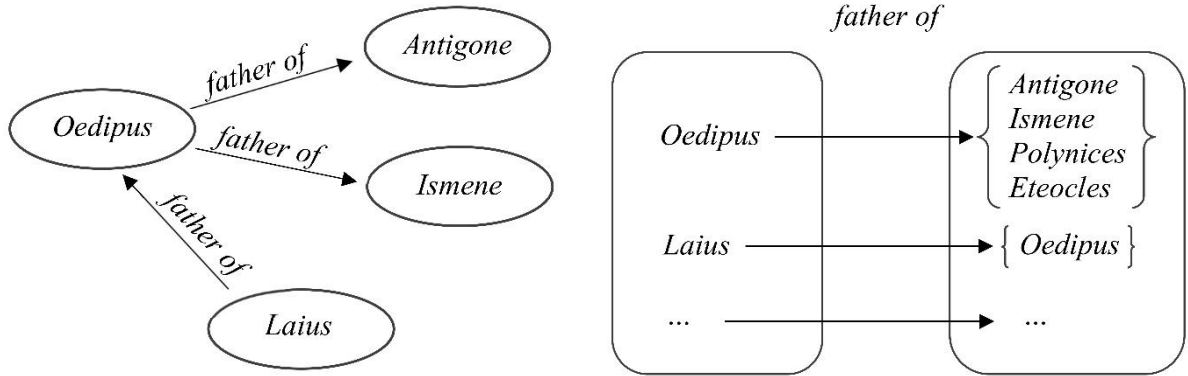

Figure 2. The semantic relation *father of*

## 4 Definitions

### 4.1 Constraints

To avoid confusion with the more ambitious notion of analogy in Hofstadter, we will use the term *constraint* to refer to abstract correspondences between two sets of data. The relation *father of* is one such constraint, taking its inputs from the domain of all the fathers and outputting, for each input, the set of his children.

For our purposes, we can consider *father of*—an elementary semantic relation—a primitive constraint so to speak, and analyse it no further. But the constraints that interest us are *composite* functions. Even a constraint as simple as the one connecting input "abc" to output "abd" is a composite of several primitive constraints, for obviously we do not keep in our memory such a relation between character-strings.

Let $f$ denote that constraint. For input "abc" to return "abd", it must travel through several simpler constraints. The first, $g$, the inverse of a simple concatenation, takes input "abc" and outputs ("ab", "c"). Then $h$, a simple relation of alphabetical succession, draws from the dataset of our knowledge of the alphabet and transforms the second element of ("ab", "c"), outputting ("ab", "d"). Finally, $l$ concatenates the two elements of its input pair and outputs "abd". Thus, $f$ is a composite of $g$, $h$, and $l$, or $f(\text{"abc"}) = l(h(g(\text{"abc"}))) = l(h(\text{"ab"}, \text{"c"})) = l(\text{"ab"}, \text{"d"}) = \text{"abd"}$. Of course, even the character-string inputs of $f$ are the outputs of a constraint that is a composite of two concatenations, one transforming "a" into "ab", and another one transforming "ab" into "abc".

It is easy to see how this can be applied to less abstract, naturally occurring analogies. For instance, a constraint at work in (3) (riddle of the Sphinx), now $f'$, takes input *morning walk* and outputs *on four legs*. If we decompose $f'$ into its primitive constituents—now using arrows for simplicity—we obtain *morning walk* ⇨ $g$ ⇨ (*morning, walk*) ⇨ $h$ ⇨ (*early stage of day, walk*) ⇨ $l$ ⇨ (*early stage, walk*) ⇨ $m$ ⇨ (*early stage of human life, walk*) ⇨ $n$ ⇨ (*baby, walk*) ⇨ $o$ ⇨ (*baby, locomotion*) ⇨ $p$ ⇨ *crawling* ⇨ $q$ ⇨ *on four legs*. This decomposition shows that $f'(\textit{morning walk}) = q(p(o(n(m(l(h(g(\textit{morning walk}))))))))) = \{\textit{on four legs}, \ldots\}$. Notice again that outputs may have more than one element, so we will say that *on four legs* $\in f'(\textit{morning walk})$ but not that $f'(\textit{morning walk}) = \textit{on four legs}$, because the simple semantic relation represented by *n,* linking *early stage of human life* to *baby*, also points, non-deterministically, to the nodes *toddler* and *child*, and similarly *p*'s output includes *going on one's belly*, *toddling*, etc.

A more complex composite constraint is the one that, for input *consequential emotional bond*, returns the set of possible one-sentence summaries of such a bond in the story of Oedipus. In its decomposition into simpler ones, this constraint will need to draw from two datasets—just as the character-string constraint had to draw from our knowledge of the alphabet—the first dataset being our knowledge of the Oedipus myth, and the second one our knowledge of how to summarise stories, so that the input undergoes the following series of simple transformations—now without function names for simplicity: *consequential emotional bond* ⇨ (*consequential relative, type of relation*) ⇨ (*Jocasta, type of relation*) ⇨ (*Jocasta, husband and wife*) ⇨ (*Jocasta, reciprocal love*) ⇨ (*Jocasta,* (*reciprocal,* "love")) ⇨ (*his mother,* (*reciprocal,* "love")) ⇨ "He loved his mother and she loved him", among other similar one-sentence summaries.

The comicality of the innocent-sounding first sentence in the output set—a line from Tom Lehrer's satirical song "Oedipus Rex" (1959)—springs from the fact that the very same words could be the output of another constraint, *differently* composed, drawing from our knowledge of a type of story poles apart from the Oedipus myth—e.g., a children's story—for the same input, *consequential emotional bond*. This duplicity is the topic of the following subsection.

## 4.2 Overlap (at constraint level)

Constraints, which have been described *intentionally* as multistep procedures transforming inputs into outputs, are of course, *extensionally*, sets of input-output pairs—not to be confused with the fact that inputs are often pairs or n-tuples in themselves. In the character-string riddle, constraint $f$ is the set {("abc", "abd"), ("ijk", "ijl"), …}. As sets, constraints intersect. Let $f'$ denote another constraint that replaces the last character in the input string with "d", regardless of what the original character is. Thus $f'$ = {("abc", "abd"), ("pq&", "pqd"), …}. We now see that the pair ("abc", "abd") belongs to both $f$ and $f'$.

Similarly, (*consequential emotional bond*, {"He loved his mother and she loved him", …}) belongs to both $f$ and $f'$, where $f$ denotes the constraint that outputs, for an input property selection, the set of one-sentence summaries of that property in an undisclosed story, and $f'$ denotes the same for the story of Oedipus.

But there is one important inaccuracy in the previous paragraph that needs addressing. In truth, the pair (*consequential emotional bond*, {"He loved his mother and she loved him", …}) does *not* belong to $f'$, for its second element is not a single sentence, but a set. Consequently, a full overlap does not occur, for this pair does not belong to $f'$—the sentence "He loved his mother with true filial love", for instance, belongs to the output set under $f$, but it does not belong to the output set under $f'$. This overlap, then, is different from the one in the character-string riddle, where the output was, deterministically, a single element, and the pair ("abc", "abd") belonged entirely to both constraints. The overlap in the Oedipus example is only *partial*, in that the outputs of $f$ and $f'$ are different, but still have elements in common, e.g., the sentence "He loved his mother and she loved him". A *partial overlap* will be equally valid for the mechanism we wish to describe.

## 4.3 Incongruity (at constraint level)

The activation of two overlapping constraints is the essence of our mechanism, but without the evocation of an incongruity the hearer would not be able to perceive the two constraints as distinct. An incongruity is an input-output pair that belongs to $f'$, but not to $f$. As should be clear by now, $f$ is the constraint that the hearer's previous experience has primed her for, whereas $f'$ is not. Thus, we can refer to *f* as the *overt constraint*, and to *f'* as the *covert constraint*. In the character-string example, the pair ("pq&", "pqd") is an incongruity. It belongs to *f'* and not to *f*.

### 4.5 Incongruity resolution

Pending the definitions in Section 5, we have the necessary constituents to describe the process of incongruity resolution. Consider, in the manner of Hofstadter and Mitchell (1995):

(7) If "abc" transforms into "abd", you surely can guess what that transformation is about. Then, what would "pq&" become if the same thing happened to it? Answer: "pq&" becomes "pqd". (Constructed example).

The unexpected input "pq&" leads the hearer into a dead end, for "&" has no alphabetical successor. The solution "pqd" sounds strange for an instant, but since her initial modelling of the situation is incompatible with the ensuing data, the hearer retraces a few steps and finds where her representation may have been too hasty. Indeed, the abstract rule constraining the transformation of "abc" into "abd" could have also been, *substitute* "d" *for the third character in the string*. Doing something analogous to "pq&" results in "pqd". The slippage—to use Hofstadter's term—from one constraint to another is the resolution of the incongruity.

Although we would lose the rhythm and elegance of Tom Lehrer's lines:

(8) He loved his mother and she loved him,
and yet his story is rather grim. (Lehrer 1959, "Oedipus Rex".)

the humour in them can be captured by a similar riddle-like text:

(9) If a *consequential emotional bond* of my story can be summarised as "He loved his mother and she loved him", you surely can guess other properties of my story. Then, how would you summarise its *pervading mood*? Answer: "His story is rather grim." (Constructed example) where the incongruous input-output pair (*pervading mood*, {"His story is rather grim", …}) forces the slippage from *f* to *f'* as these two constraints were described in Section 4.2.

## 5 Mechanism of verbal humour

Does the above description account for all of humour? Hofstadter has not developed his notion of slippage into a general-purpose theory of humour, but he famously pointed out the similarity between the formal structure of the character-string riddle and this joke from the Cold War era:

(10) An American says: Look how free we are in America—nobody prevents us from parading in front of the White House and yelling, "Down with Reagan!" And a Russian replies: We in Russia are as free as you—nobody prevents us from parading in front of the Kremlin and yelling, "Down with Reagan!" (Quoted in Hofstadter and Gabora 1989: 427)

Although it would not be the most effective way to tell it, the joke can be rephrased as a riddle whose form is almost identical to (7): If America is a free country because input *America* transforms into the true sentence, "Nobody prevents Americans from parading in front of the White House and yelling 'Down with Reagan!'", would input *Russia* give us a true sentence if the same thing happened to it? Yes, it would, etc., where the literal insertion of "Reagan" at the end of the output for *Russia* is analogous to the literal insertion of "d" at the end of the output for "pq&"—both are constant values instead of the expected input-driven values.

Obviously, the character-string riddle seems too abstract to elicit so much as a polite smile in most contexts, and as for the Cold War joke, it is unusual in that few jokes can be described so straightforwardly by formal operations on strings of symbols. But we believe that two simple additions will give us a truly generalisable mechanism of humour.

### 5.1 The narrative joke

We will now arrange the elements defined in Section 4 into a mechanism capable of describing the processing of a more standard narrative joke:

(11) A man with a hunchback gets lost in a cemetery at night. A ghost spooks him. "What's that on your back?" the ghost asks. "It's a hunch," the man replies. "Can I have it?" the ghost asks. "Sure!" And the trade is done. Overjoyed, the man goes to his limping friend and tells him the good news. The limping man goes to the cemetery right away and starts wimping about looking for the ghost. The ghost shows up. "What's that on your back?" the ghost asks. "Er… nothing," the man replies. "Here, have this hunchback!" (Traditional joke)

In (7) ("abc" riddle) and (10) (Reagan joke), constraint outputs were sentences or symbol-strings included verbatim in the final utterances. A narrative joke does not normally work that way, (11) being a good example. We have moved from the formal manipulation of strings of symbols, so easy to represent formally, to the realm of fiction. Ritchie (2018) and Attardo (2001), noting that most jokes are in essence very short "stories", have applied the tools of fictional-text comprehension theories to the analysis of humour in jokes and longer narratives. Promising though this is, we will take the opposite approach: since all jokes are also, by definition, simply texts, we will restrict ourselves to the assumptions about communication stated in Section 3 and the definitions in Section 4. This eliminates the complication that some humorous texts are non-fictional—conversational humour, witticisms such as (24)—and the even greater one that "there is no consensus on how to represent, formally, the content of a story" (Binsted and Ritchie 1996). Most crucially for our goals, it will prevent us from introducing in our general-purpose mechanism any element not to be found in non-verbal communication.

Assuming that analogies are constantly being evoked in human communication, including that of fictional content, all we need to represent formally are such analogies, and we need not worry about other constituents of the story, important as they may be from a narratological perspective. We call these analogies *narrative constraints* only because they play a role in the narrative, but we do not need to overcomplicate our task by analysing what different narrative functions they may play in the storyworld.

Thus, as one represents mentally the meaning of the setup of (11), the constraint in Figure 3 is evoked.

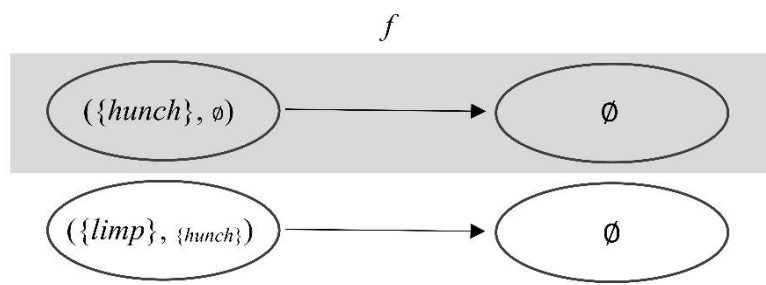

Figure 3. Overt narrative constraint (*f*) in the hunchback's joke

Input ({*hunch*}, ∅) in the upper row represents the state of affairs when the hunchback encounters the ghost. The first element in the pair, {*hunch*}, is the set of the hunchback's physical defects. The second element is the set of known defects kept by the ghost, at this moment the empty set, denoted ∅. This second element is actually irrelevant for constraint *f*, which simply ignores it—the reason for its inclusion in the input will become apparent below. The output, ∅, is the set of the hunchback's defects after the trade is done. Informally described, *f* is simply *the ghost removing any element she finds in the set of defects of the input character*.

*f* is not explicit in the utterances, of course. It is not even one of its implicated propositions in the sense linguists normally give to *implicature*. Its input and output are indeed propositions implicated by the text, but not literal parts of it like the symbol-strings of our previous examples. In sum, *f* is only a relationship between data that gets evoked spontaneously in the hearer's cognitive representation of the discourse, without any conscious effort, illustrating the way the mind routinely distils from individual data more abstract, generalisable descriptions.

When the limping man encounters the ghost, the hearer expects input ({*limp*}, {hunch}) to return ∅, its output under *f* (Figure 3, bottom row), but what happens instead is what is shown in the bottom row of Figure 4.

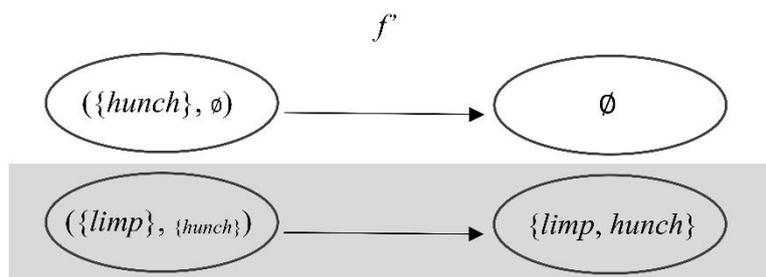

Figure 4. Covert narrative constraint (*f'*) in the hunchback's joke

In this new representation, the ghost does not go about curing the defects of those she encounters. She is only interested in hunches, hence she asks, "What's that on your back?" to the limping man, to his bewilderment. What she seems to be in the habit of doing is *f'*, *swapping between passers-by the property of having or not having a hunch*. That is why she took the hunch from the hunchback, but the *analogous* thing she does when she meets someone who does not have a hunch and she keeps one is giving it to him.

## 5.2 Overlap (at utterance level)

As these outputs are not explicitly included in the final utterances, we will define a new overlap at the level of utterances—different from, but resulting from, the overlap at constraint level. The new overlap is the set of utterances that can implicate both the application of *f* and the application of *f'* to the input datum, as seen in Figure 5. An input value returns an output under each constraint, and for each output there is a corresponding set of all possible utterances implicating it. These two sets overlap when there is a partial or full overlap in the abstract realm of the constraints' outputs, replicating it in the concrete domain of the final stimuli implicating such outputs. This overlap is then the set of utterances capable of evoking the cognitive representation of both constraints, because the outputs of both can be inferred as implicatures from such utterances.

For input ({*hunch*}, ∅), *f* and *f'* overlap fully, returning the same output, ∅—highlighted in grey. This overlap at constraint level causes a large overlap in the sets of all possible utterances implicating one and the other constraint, i.e., the two circles $F_0$ and $F'_0$ almost coincide. The utterances describing the hunchback's good fortune belong to that intersection.

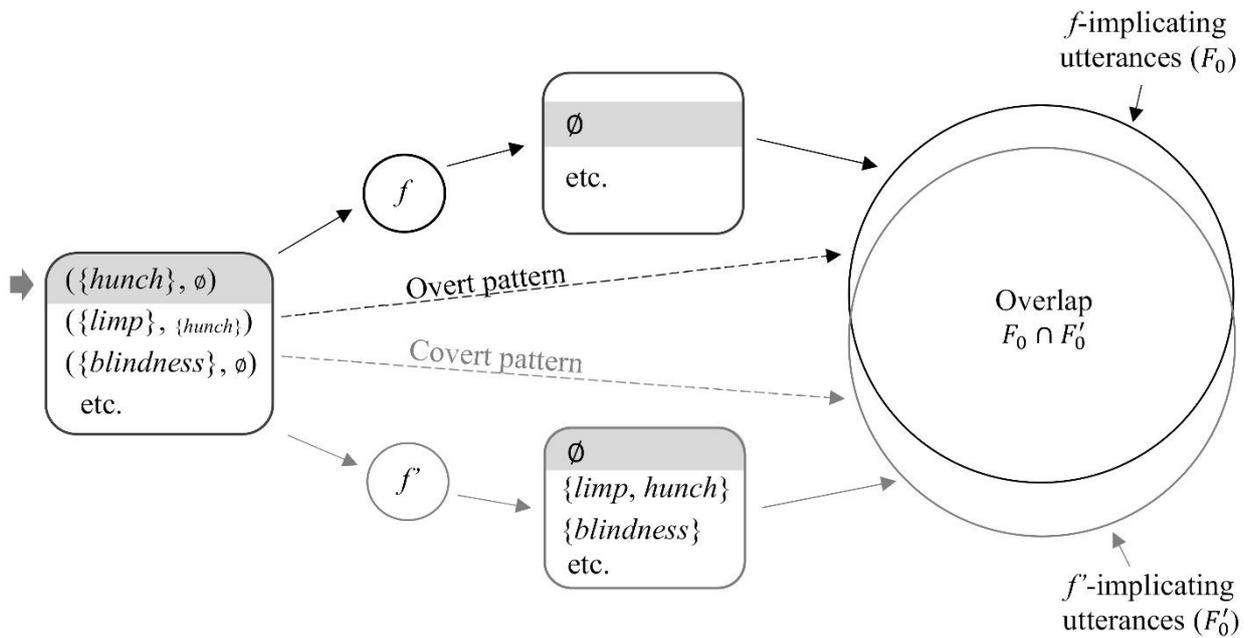

Figure 5. General mechanism in the hunchback's joke, for input ({*hunch*}, ∅), highlighted

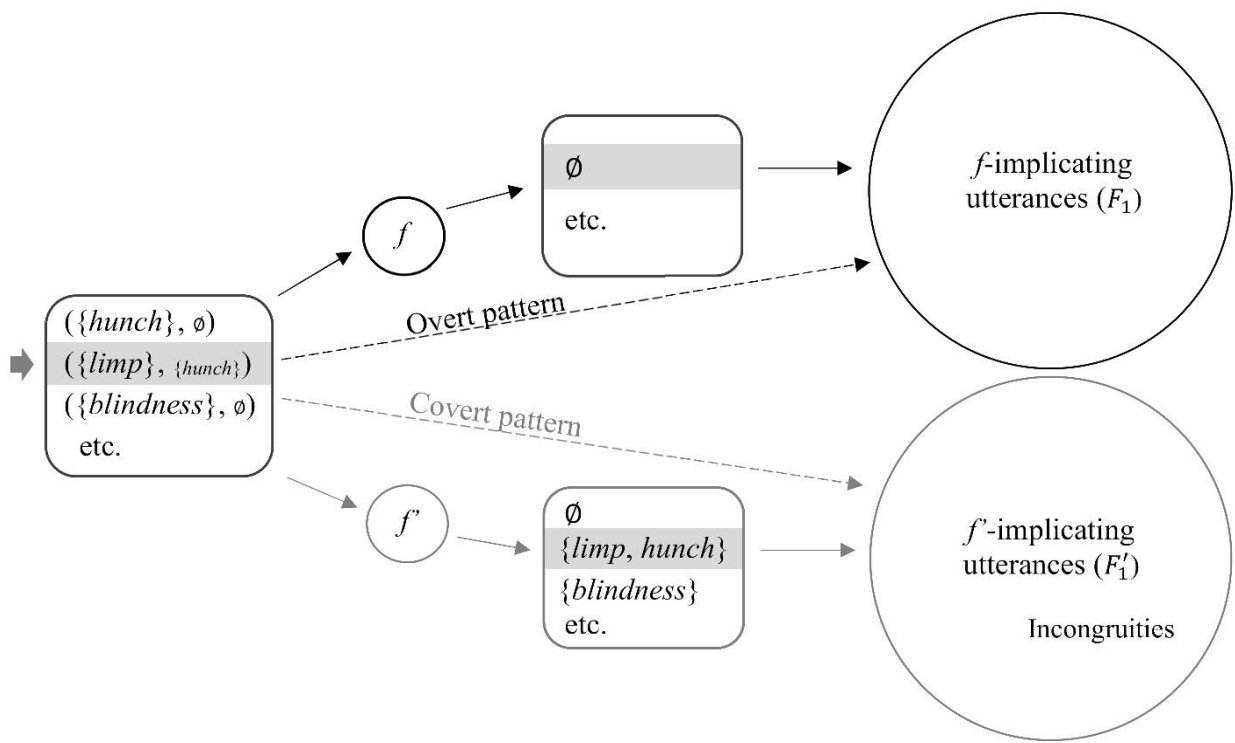

Figure 6. General mechanism in the hunchback's joke, for input ({*limp*}, {*hunch*}), highlighted

But for input ($\{limp\}$, $_{\{hunch\}}$) the outputs of $f$ and $f'$ are different, as shown in Figure 6. There being no constraint overlap, the sets of possible utterances implicating one and the other output, $F_1$ and $F'_1$, do not intersect, and the utterances describing the limping man's misfortune belong to $F'_1$ only, being perceived as incongruities through the lens of the previous pattern.

The same stream of stimuli—in this case, the utterances narrating the hunchback's episode—can often serve as evidence from which to infer more than one narrative constraint. So it is often the case that two different constraints *overlap*, that is, are inferable from the same set of utterances—or, as we will see later, stimuli, more broadly considered—without the hearer noticing that she is paying attention to only one of them, until the input switches—to the limping man's initial state of affairs—and returns an incongruity.

### 5.3 Incongruity (at utterance level)

The *incongruity at utterance level* is an utterance in the set of those that implicate the covert constraint's output, which does not belong to the set of those that implicate the overt constraint's output. Thus, the incongruity exposes the overlap, for without it the hearer would not evoke $f'$ and would—at most—evoke only $f$. It is the incongruity that compels her to adopt $f'$ in her cognitive representation, which is what we call getting the joke, or the "Aha!" moment.

Unsurprisingly, this process is inaccessible to the hearer's introspection, who gets the joke in a fraction of a second, just as speakers of natural languages produce sentences the syntactic structure of which they are completely unaware of—unless they happen to be doing a syntax exercise—an elementary matter of fact of linguistics and cognitive science.

### 5.4 Overt pattern and covert pattern

For the sake of brevity, hereinafter we will refer to the overt constraint as $f$, and the covert constraint as $f'$. Their full overlap will be denoted $f \cap f'$, and their partial overlap $f \cap^p f'$. Similarly, we will refer to the set of utterances from which the output of $f$ for the first input can be inferred as an implicature as $F_0$, and the set of utterances from which the output of $f'$ can be inferred as $F'_0$, so that the overlap at utterance level is $F_0 \cap F'_0$. The sets of utterances implicating the outputs of $f$ and $f'$ for the second input will be denoted $F_1$ and $F'_1$. The relationship between the inputs and the $F$ sets is the *overt pattern*, and that between the inputs and the $F'$ sets is the *covert pattern*.

### 5.5 Setup and punchline

We have been slow to give an explicit definition of *setup* and *punchline,* the reason being that, although useful concepts when analysing canned jokes, they do not have a fixed place in the mechanism of Figures 5 and 6. Thus, (1) and (2) (lightbulb joke and its musical version) are described by the same mechanism regardless of the fact that the punchline of (1) is the setup of (2), and vice versa. The same happens in (3) and (4) (riddle of the Sphinx and its witticism version). Our mechanism describes the processing of sets of stimuli without differentiating between setup and punchline.

However, in the particular case of a canned joke, utterances unfold sequentially in time, and the *punchline* can easily be defined as the last utterance that completes the minimum set of utterances needed for the hearer to evoke the overlap. In a typical canned joke, the punchline achieves this by communicating the incongruity—"Here, have this hunchback!"—but this need not always be the case (see Section 5.10). The *setup* can then be defined as the sequence of utterances that come before the punchline.

In humour that does not unfold sequentially—cartoons, slapstick—this distinction is not likely to be useful.

## 5.6 The overlooked argument

Finally, this important element can be properly introduced. We do not ignore that the mechanism outlined above is still incomplete, for many instances of non-humorous disambiguation fit that description, as pointed out by Ritchie:

(12) A man jumped into a taxi and said, "The main railway station, please." "OK," said the driver, "What time is the train?" "No train," said the passenger, "I'm meeting a friend in the bar there." (Ritchie 2018: 95)

Similarly, (11) would not carry humour if the ghost, instead of putting the hunch on the limping man, simply replied to him, "Sorry, I don't take limps."

What is important to notice is that *f* is typically a probability distribution to the effect that the taxi passenger is *likely* going to take a train, and the ghost is *likely* to take the limp, based on the hearer's previous knowledge and the available information. But information is always incomplete, and re-evaluating the likelihood of upcoming data as new information becomes available is *not* what makes *f'* different from *f*.

To use a graphic analogy, the grey stripes highlighting the outputs in Figures 5 and 6 are a simplification, for in many cases they should be thought of as gradients from grey to white, encompassing several outputs from the most likely—grey—to the least—white (Figure 7). Similarly, the circles *F* and *F'* should be thought of as spotlights with a blurry penumbra around the edges, the most likely utterances occupying the fully lit centre, and the least likely far out. In (12), the passenger's final reply would still get some dim light, so it would not be incongruous. No new constraint *f'* is evoked, and *f* remains a valid representation, even if the updated context may in turn serve as a basis for the evocation of new constraints with which to anticipate upcoming portions of the discourse.

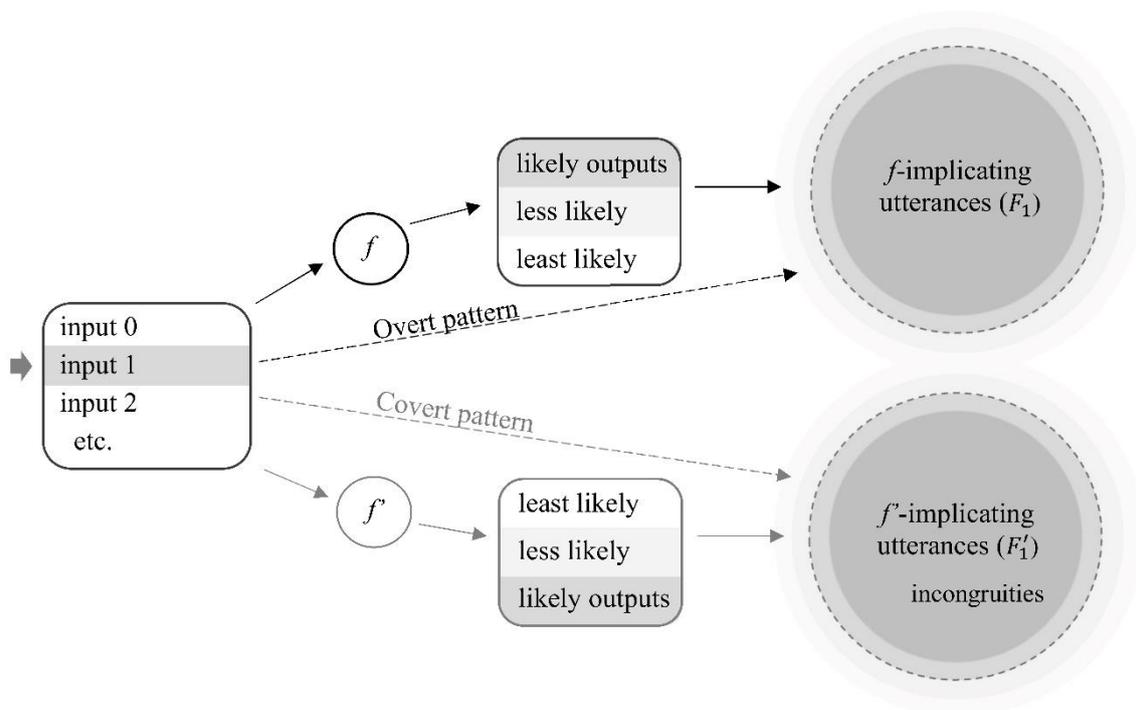

Figure 7. *f* and *f'* represented as probability distributions

Figure 8 shows how this is equivalently expressed by saying that the input of *f'* takes one additional argument *overlooked* by *f*, and whose value *is* implicit in the available information. Thus, in (11) *f* ignores the second element of the input pair ({*limp*}, $_{\{hunch\}}$), written in $_{subscript}$ for easy identification. The overlooked argument is the set of defects kept by the ghost, which has value $_{\{hunch\}}$ when the limping man meets her. Unlike *f*, *f'* factors it in. In (3) and (4) (riddle of the Sphinx), the overlooked argument is the timespan that the expressions "morning", "noon", and "evening" subdivide, whose value, $_{life}$, is factored in by *f'* whereas *f* simply assumes the twenty-four-hour period the semantic field of "morning", "noon", and "evening" primes the hearer for. In (7) ("abc" riddle), the overlooked argument is the character to substitute for the last one in the string, taking value $_{\text{"d"}}$, and similarly in (10) (Reagan joke), taking value $_{\text{"Down with Reagan!"}}$. In both cases, *f* ignores this argument under the assumption that the value to substitute is driven by the value of the other, non-overlooked argument in the input. In (8) (Oedipus song), the overlooked argument is the specific character from which to draw the data, taking value $_{Oedipus}$, and which *f* ignores under the assumption of a more ordinary undisclosed story evoked by the semantic fields of "mother" and "love".

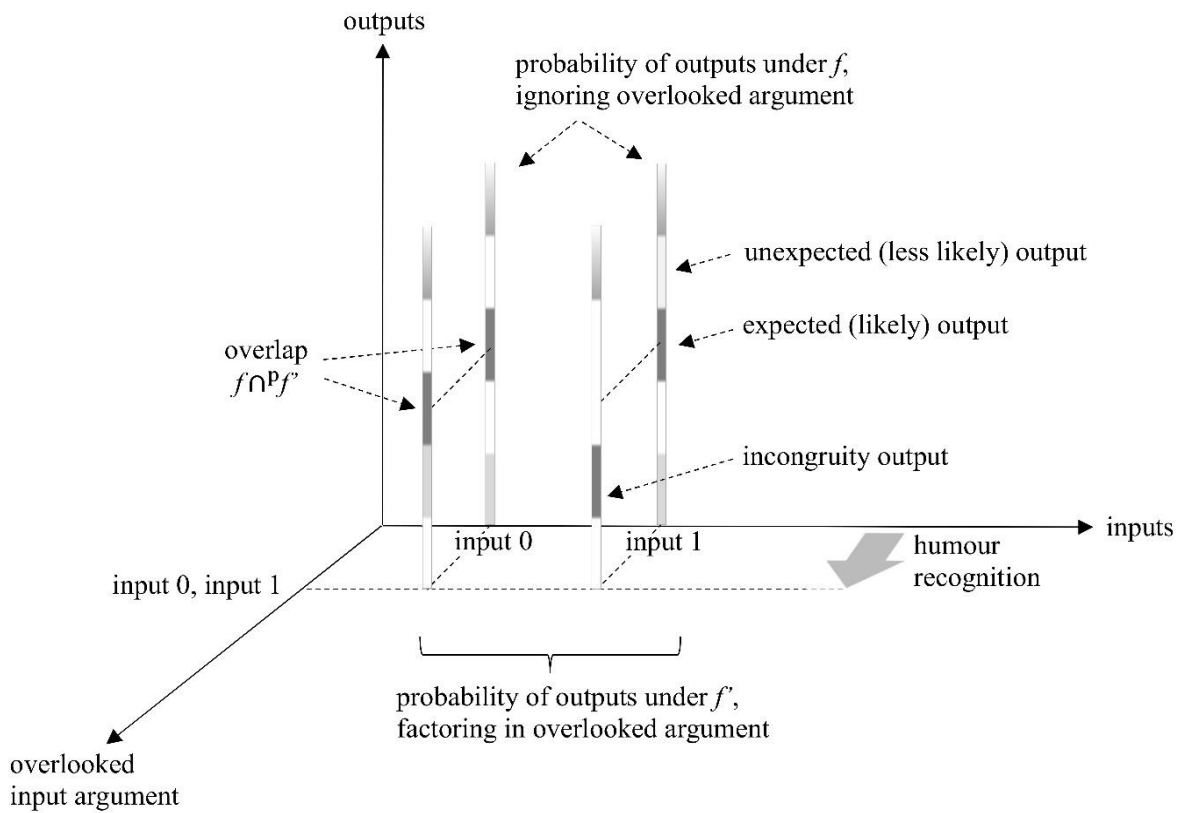

Figure 8. The overlooked argument represented as a new dimension of the input

Considering *f* and *f'* probability distributions makes most—if not all—constraint overlaps partial. For expository clarity, though, we will ignore these subtleties and continue treating constraint outputs as all-or-nothing sets, and *F* and *F'* as clear-cut circles. This brief incursion into probability was needed to clarify that the incongruity is *not* the unlikelihood of a stimulus in the output set, but the "shift forward" to an entirely different probability distribution output on a new plane (Figure 8). This also shows the difference between our model and the "debugging" theory of humour (Hurley et al. 2011). It is not just *any* kind of belief or idea that, having surreptitiously entered the mental space built by the hearer, is diagnosed to be false, but a particular kind of idea, a correspondence between data, that is diagnosed to be, not necessarily false, but one dimension short of modelling the discourse situation correctly, i.e., flat when the input takes a second argument, etc.

This *factoring in* of an additional argument makes *f'* a more "sophisticated" model of the situation than *f*, accounting for the hearer's subjective feeling of having been "outsmarted" by the joke teller, the feeling of "getting it", and the spatial metaphors of change of "angle" and "point of view" used by comedy writers, reflecting the fact that the overlooked argument is a normally unseen dimension along which the input vector can nonetheless vary. The overlooked argument is essentially what makes the constraint overlap a true overlap, and not a mere update of the context as new information comes in. Its covert, subjacent character gives ground for its frequent association with a few favourite inhibited topics like sex and disparaging stereotypes. In Raskin's terms, the overlooked argument accounts for the local opposition of scripts.

### 5.7 Script Opposition

As it would be trivial and lengthy to reanalyse the previous examples in this light, we will move on to show how the mechanism in Figures 5 and 6 accounts for the phenomenon of script overlap and opposition, at the core of the SSTH/GTVH, using the exemplar text in the linguistics of humour, the doctor's wife joke famously analysed by Raskin:

> (13) "Is the doctor at home?" the patient asked in his bronchial whisper. "No," the doctor's young and pretty wife whispered in reply. "Come right in." (Quoted in Raskin 1985: 100).

In Raskin's semantic theory, the setup of a joke is compatible with the activation of two opposed scripts. In the terms of Figures 5 and 6, *f* takes as input an element from the set {*assessment*, *resolution*} and outputs the set of speech acts necessary to ascertain a doctor's presence/absence—for input *assessment*—and the set of speech acts resulting from his presence/absence—for input *resolution*. The input-output pair (*assessment*, {*asking if the doctor is at home*, …}) also belongs to $f \cap^p f'$, where *f'* takes input (*assessment*, <sub>sexual encounter</sub>), whose second argument—the purpose of the visit—is overlooked by *f*, for *f* assumes the medical purpose the hearer is primed for. The value it takes, <sub>sexual encounter</sub>, is made unmissable enough for some audiences by the whispering, and by the wife being described as young and pretty. For that input, *f'* outputs the set of speech acts necessary to ascertain, not the doctor's, but the *husband's* presence/absence, some of which happen to be in the previous set too—as is the case of *asking if the doctor is at home*—hence the partial constraint overlap. This overlap causes $F_0$ and $F'_0$, the sets of utterances that perform those acts, to overlap too. The utterances in the joke's setup belong to $F_0 \cap F'_0$. But the last utterance, her invitation to come in, belongs only to $F'_1$, for it results from the pair ((*resolution*, <sub>sexual encounter</sub>), {*inviting lover in*, …} or the set of speech acts resulting from husband's absence), which belongs to *f'* but not to *f*. There are diverse ways to model this cognitive process, including of course Raskin's, but all of them are, in essence, equivalent.

In fact, if we assume that semantic script activation is a valid model of text processing, constraint *f* cannot fail to activate the script for MEDICAL VISIT, and constraint *f'* the script for ADULTERY, since each of these two scripts accommodates the semantic relations that the corresponding constraint draws from. Given that the notion of *semantic script* as a "cluster of nodes and their connections in the semantic network" is broad enough to encompass a wide variety of "chunks" of information, that scripts can be dynamically updated as the hearer accumulates new information, and that they "can be activated without a lexematic handle, i.e., inferentially" (Attardo 2020: 116-126), then a necessary outcome of a constraint overlap is the overlap of two opposing scripts of one kind or another. In (11) (the hunchback's joke) the opposing scripts would be the script for AFFLICTION REMOVAL—familiar from miracle and supernatural stories—and the script for SWAPPING CARRIER OF PROPERTY—a more general and abstract script. The only limit is what counts and does not count as a semantic script—but a theory of humour capable of handling non-verbal humour should contemplate scripts well beyond the lexicon—and what counts and does not count as Script Opposition. In (10) (Cold War joke), an explanation in terms of ready-made knowledge packages that stand in opposition to each other becomes more complicated. Section 5.9 will present a case that is difficult to analyse in script theory. Our description in terms of overlapping constraints makes these cases easier to handle, and we hope to show it is more easily generalisable to non-verbal humour.

## 5.8 Linguistic humour

In (13), the outputs of *f* and *f'* are sets of speech acts, not the sets of physical defects with which we represented the hunchback's joke. It is essential to this model that the constraints' outputs may be *any datum* that can be cognitively represented, from character strings to sets of physical defects, speech acts, words, sentences… This affords us an elegant way to characterise the old distinction (Attardo 1994: 27) between playing with ideas and situations—situational humour— and playing with words—linguistic humour, including puns. In linguistic humour, the outputs of constraints *f* and *f'* happen to be linguistic signifiers—words or sentences—or sets of them:

(14) "Every forty-five seconds someone in the US sustains a head injury."
"Omigod! That poor guy!" (Contemporary joke)

In this case, *f* takes its inputs from the set {*reporting*, *reacting to*}, and outputs the set of sentences that report/react to the proposition, "∀ 45-second interval, ∃ a US citizen that sustains a head injury in that interval." The input-output pair (*reporting*, {"Every forty-five seconds someone in the US sustains a head injury", …}) belongs to $f \cap^p f'$ too, where *f'* takes input (*reporting*, *a US resident x*) and outputs the set of sentences that report the proposition that *x* sustains a head injury every forty-five seconds. The overlap is partial, for one can think of sentences that belong to *f*'s output and not to that of *f'*, such us, "Every forty-five seconds someone *different* in the US sustains a head injury". The second argument in the input, taking value *a US resident x*, is overlooked by *f*, because the sentence primes the hearer for a statistical interpretation and the appropriate quantifier order. The partial overlap results in an overlap at utterance level, $F_0 \cap F_0'$, to which the setup belongs. When the input changes to (*reacting to*, *a US resident x*), the outputs of *f* and *f'* do not overlap, and the punchline belongs to $F_1'$ and not to $F_1$, revealing the overlapping *f'*.

(14) is obviously not a proper pun, a topic left for Section 5.10, but it illustrates the observation that, "from a very broad viewpoint, most humorous texts can actually be considered to contain a pun: the pun is that part of the humorous text, possibly an implied and not overtly expressed part, that is compatible with two meanings", with the provision that "the meaning of a text is not simply the combination of the meanings of its individual words" (Hempelmann and Miller 2017: 95). Our model's "rich" sense of *meaning* involves the evocation of constraints. When the "polysemous" output of two overlapping constraints is a word, we speak of puns. When it is an implied datum instead of a surface item, as in (11) (the hunchback's joke), we speak of situational humour. (14) is an intermediate case, in which the output is a string of words.

## 5.9 Problem cases for the SSTH/GTVH

Although the primary goal of our reformulation of the semantic mechanism of humour in terms of constraints is its application to non-verbal humour, it is possible that our mechanism's general-purpose character be useful in the analysis of jokes that a script-based model may have difficulties with. Raskin's account of the following verbalization of a Stan Hunt's cartoon in terms of "rather sophisticated" semantic scripts is admittedly cumbersome and tentative:

(15) A man sitting in his living room in front of the tv set turns to his wife and says, "Funding for the *Dick Cavett Show* has been provided by this station and other public-television stations, and by a grant from the Chubb Group of Insurance Companies, with additional funding from Allen Services Corporation." (Hunt, quoted in Raskin 1985: 137-138).

But the joke can easily be mapped onto the mechanism of Figures 5 and 6. For an input taken from the set {*level of interest*, *literal content*}, $f$ outputs the value of the input property selection for a customary piece of information on tv. The input-output pair (*level of interest*, {*interesting enough to relay to wife*}) belongs to $f$, but also to $f \cap f'$, where $f'$ takes input (*level of interest*, display prominence) and outputs whether the Dick Cavett Show acknowledgement of funding is worth relaying based solely on the prominence with which it is displayed—an argument overlooked by $f$—instead of its intrinsic informative value. For the second input, *literal content*, the outputs differ. The output expected under $f$ is a customary piece of news that is intrinsically interesting enough to be relayed to wife, whereas $f'$ outputs the literal quotation of a prominently displayed acknowledgement of funding. The husband's turning to his wife to relay that information belongs to the overlap $F_0 \cap F_0'$, but his actual utterance belongs to $F_1'$ and not to $F_1$.

Interestingly, $f$ is not evoked by the hearer before $f'$ is, and this marks an important departure from (11) (the hunchback's joke), in which the input-output pair (({*hunch*}, ∅), ∅) could be said to be noticed by the hearer *before* the second input-output pair (({*limp*}, {hunch}), {*limp*, *hunch*}) was noticed, forcing the slippage from $f$ to $f'$. In (15), the input-output pairs (*level of interest*, {*interesting enough to relay to wife*}) and (*literal content*, {"Funding for the Dick Cavett Show…"}) are processed simultaneously by the hearer. We will go over this important property of the mechanism in the following sections.

## 5.10 Problem cases. Transposed incongruity

The same flexibility in the linear organisation of the constitutive elements can be observed when a joke's punchline, typically a pun, is not incongruous, as Attardo has noticed:

(16) Bassompierre was a prisoner at the Bastille. While reading, he flipped the pages of his book hastily. The warden asked him what he was looking for, and Bassompierre replied: "I am looking for a passage, but I cannot find it." (Quoted in Attardo 2020: 192)

(17) Two atoms are walking down the street and they run into each other. One says to the other, "Are you alright?" "No, I lost an electron!" "Are you sure?" "I'm positive!" (Quoted in Yus 2016: 144)

The punchline of (16) is congruent unless the hearer collaboratively engages in interpreting it so as to make it incompatible with the setup. And the punchline of (17) is congruent in either of its two possible interpretations. Our mechanism provides a simple explanation. It makes more sense to place both punchlines in $F_0 \cap F_0'$, the overlap. In (17), $f$ takes inputs from the set {*lexical labels of state*, *behaviour*} and outputs the value of the input selection for an atom missing one electron. The input-output pair (*lexical labels of state*, {"positive", …}) belongs to $f \cap^p f'$, where $f'$ takes input (*lexical labels of state*, prosopopoeia) and outputs the set of lexical labels {"positive", "unsure", …}, treating the atom in the fictional manner indicated by the second argument, overlooked by $f$. The signifier "positive" belongs to both outputs—albeit with different meanings. But for input (*behaviour*, prosopopoeia), $f'$ outputs a very different set of behaviours, including the atom's talking and walking down the street, and this is the incongruity.

Interestingly, the incongruity comes first, in the setup, and the overlap last, in the punchline. Perhaps for that reason, the covert and the overt pattern exchange roles, for the setup primes the hearer for constraint *f'*, which can rightly be called the "overt" one, whereas *f* is not as easily accessible. Bassompierre's joke would get an identical analysis. The resemblance of these two jokes to the subgenre Binsted and Ritchie (1996) have called *story puns* suggests to us that a similar description could be made of these.

Once again, this analysis assumes that the way the overlap and the incongruity are related to one another is logical, not temporal. The processing of a humorous communicative event, which comes down to the perception of the overlap, is not restricted to a specific order of presentation or processing of its constituent elements. This observation, which is in no way original (Oring 2003: 2) and has been emphasised by researchers in relation to visual genres, such as newspaper cartoons and advertising (Dynel 2008, Hempelmann and Samson 2008), will be essential to our model's account of non-verbal humour.

# 6 Humour in literature and drama

## 6.1 Stage and screen humour

The mechanism accounts similarly for humour in genres other than the canned joke and the riddle. In the following joke, so characteristic of *King Lear*, the Earl of Kent has been admitted into Lear's service as a trusted counsellor just as the first signs of Lear's downfall become apparent.

(18) LEAR   Now, my friendly knave, I thank thee; there's earnest of thy service.
          [Gives KENT money]
   Enter FOOL
   FOOL   Let me hire him too; here's my coxcomb. [Offers KENT his fool's cap.]
   LEAR   How now, my pretty knave, how dost thou?
   FOOL   [To KENT] Sirrah, you were best take my coxcomb.
   KENT   Why, my boy?
   FOOL   Why? For taking one's part that's out of favor.
(Shakespeare, *King Lear*, 1.4)

When the Fool says, "Let me hire him too", he claims to be hiring Kent in the same manner as Lear, forcing a reinterpretation of the contractual relationship between Kent and Lear. For input *affiliation*, *f* outputs {*hired*}. But the input-output pair (*affiliation*, {*hired*}) belongs to *f*∩*f'*, where *f'* takes input (*affiliation*, <sub>out of favor</sub>) and thus it factors in Lear's political standing, overlooked by *f*. For the second input (*capacity*, <sub>out of favor</sub>), *f'* outputs {*fool*, …} as the capacity Kent is hired in, which is implicated by the Fool offering Kent his coxcomb and hiring him as well—stimuli that belong to $F'_1$—and further explained by the reasoning that only a fool would enter the service of a downfallen political figure. The output {*fool*, …} is incongruous under *f*, which outputs {*counsellor*, …} for the second input, thus revealing the covert pattern overlapping Lear's actions and words, such as thanking and paying Kent—stimuli that belong to $F_0 \cap F'_0$.

Running gags and similar techniques illustrate the preference for the overlaps of successive jokes to be themselves constrained by a shared mechanism—disconnected jokes are not the most successful kind of longer-format comedy writing. An extended excerpt from a Monty Python sketch will illustrate this. The scene is "a sitting room straight out of D. H. Lawrence. Mum, wiping her hands on her apron, is ushering in a young man in a suit. They are a Northern couple."

(19) MUM      Oh dad ... look who's come to see us ... it's our Ken.
    DAD      *(without looking up)* Aye, and about bloody time if you ask me.
    KEN      Aren't you pleased to see me, father?
    MUM      *(squeezing his arm reassuringly)* Of course he's pleased to see you, Ken, he ...
    DAD      All right, woman, all right I've got a tongue in my head—I'll do t'talkin'. *(looks at Ken distastefully)* Aye ... I like yer fancy suit. Is that what they're wearing up in Yorkshire now?
    KEN      It's just an ordinary suit, father ... it's all I've got apart from the overalls.
*Dad turns away with an expression of scornful disgust.*
    MUM      How are you liking it down the mine, Ken?
    KEN      Oh it's not too bad, mum ... we're using some new tungsten carbide drills for the preliminary coal-face scouring operations.
    MUM      Oh that sounds nice, dear ...
    DAD      Tungsten carbide drills! What the bloody hell's tungsten carbide drills?
    KEN      It's something we use in coal-mining, father.
    DAD      *(mimicking)* 'It's something we use in coal mining, father.' You're all bloody fancy talk since you left London.
    KEN      Oh not that again.
    MUM      *(to Ken)* He's had a hard day dear ... his new play opens at the National Theatre tomorrow.
    KEN      Oh that's good.
    DAD      Good! *good?* What do you know about it? What do you know about getting up at five o'clock in t'morning to fly to Paris ... back at the Old Vic for drinks at twelve, sweating the day through press interviews, television interviews and getting back here at ten to wrestle with the problem of a homosexual nymphomaniac drug-addict involved in the ritual murder of a well known Scottish footballer. That's a full working day, lad, and don't you forget it!
    MUM      Oh, don't shout at the boy, father.
    DAD      Aye, 'ampstead wasn't good enough for you, was it? ... you had to go poncing off to Barnsley, you and yer coal-mining friends. *(spits)*
    KEN      Coal-mining is a wonderful thing father, but it's something you'll never understand. Just look at you!
    MUM      Oh Ken! Be careful! You know what he's like after a few novels.
(*The Monty Python's Flying Circus*, 1969, Ep. 2)

All the information needed to represent the overlapping patterns is easily accessible in the audience's shared background knowledge, obtained from previous exposure to this kind of drama. Being a long text, it activates several constraints simultaneously—or a single constraint with many variables, depending on how one wishes to look at it. For the sake of simplicity, let *f* take as inputs elements from the set {*types of attitude*, *targets of attitude*}, and output the value of the input property selection for the father-character in this stereotypical, recognisable scene. The input-output pair (*types of attitude*, {*resentment*, *antipathy*, …}), belongs to *f*∩*f'* too, where *f'* takes input (*types of attitude*, (playwright, coalminer)), and thus it factors in which occupations correspond to father and son, a datum overlooked by *f*, which assumes the familiar working-class father and artistically ambitious son, the reverse of what *f'* contemplates. Utterances and stage directions implicating output {*resentment, antipathy*, …} are compatible with both patterns, but when the input changes to (*targets of attitude,* (playwright, coalminer)), *f'* outputs targets such us {*living in Yorkshire*, *using coalmining words*, *being condescending towards the achievements of an intellectual*, …}, leading to utterances that are incongruous under the overt pattern, such as the father saying, "Tungsten carbide drills! What the bloody hell's tungsten carbide drills?" instead of the expected, "*Esprit du temps!* What the bloody hell's *esprit du temps!*" Other instances of role-reversal humour could be analysed similarly.

## 6.2 Humorous plots

Attardo (2001: 92-98) makes the important distinction between humorous narratives with a humorous plot or fabula, and humorous narratives with an essentially serious one. Shakespeare's *As You Like It* may contain much humour, but its humour can be said to be superimposed on an otherwise non-humorous, romance-like fabula. On the other hand, the plot of Poe's *The System of Dr. Tarr and Dr. Fethers* is structurally analogous to a joke in that it ends in a punchline, forcing a reinterpretation of the entire text. We also encounter *plots with humorous fabulae*, which do not necessarily end in a punchline, but have central narrative complications involving humour, such as Wilde's *Lord Arthur Savile's Crime*. The story begins when Lord Arthur has his palm read by a chiromancer, who tells him that it is his destiny to become a murderer. Since he is soon to be married, Lord Arthur realises that it is his moral duty to commit the murder—unavoidable after all—*before* marrying. Attardo points out that this realisation, central to the story, bears an unmistakable resemblance to a punchline, and calls such an element a *jab line*. Unlike punchlines, jab lines are not detachable from the overall narrative to constitute stand-alone jokes by themselves, because their humour relies on the narrative context. They do not disrupt the narrative flow, and may even be essential elements in the development of the plot.

Mapping Lord Arthur's central complication onto our mechanism can easily be done if we consider the two overlapping constraints evoked by the following excerpt:

(20) He recognised none the less clearly where his duty lay, and was fully conscious of the fact that he had no right to marry until he had committed the murder. (Wilde, *Lord Arthur Savile's Crime*, Ch. III).

For input *marriage*, $f$ outputs Lord Arthur's judgement on the morality of carrying out that deed before committing murder. Thus, the input-output pair (*marriage*, {*NO right to*}) belongs to $f$. But it also belongs to $f \cap f'$, where $f'$ takes input (*marriage*, {*marriage, murder*}), factoring in, in addition, a set of two acts, and outputs his judgement on the morality of choosing to marry first, when the only choice at his disposal is the *when*, and not the *what* is to be done—the two acts being presumed unavoidable. This shared input-output pair results in the overlap $F_0 \cap F'_0$, to which the sentence "he had no right to marry until he had committed the murder" belongs. But for input (*murder*, {*marriage, murder*}), $f'$ necessarily outputs {*moral duty to*}, which is logically entailed by the output of the previous input, whereas $f$ still outputs {*NO right to*}. From this it follows the incongruity of Lord Arthur's realisation that it is his moral duty to commit murder as soon as possible if he must. The literary articulation of this thought is in $F'_1$, the set of possible verbal stimuli resulting from the covert pattern, but not in $F_1$, the output set of the overt pattern.

It is little surprise that some humorous longer texts feature, as key constituents of the plot, events that activate an incongruity-resolution process in themselves, even to the point of giving the overall narrative a joke-like structure. The plot of the film *Mars Attacks!* (1996) can be summarised as a sort of joke: When everything else fails, what lethal weapon can we earthlings engineer to repel a Martian invasion? The music of Slim Whitman. Incidentally, Aristophanes' *Lysistrata* would turn into a somewhat similar joke.

We see no reason to think that other properties of longer humorous texts pointed out by recent scholarship, such as *shadow opposition*, *dissipated trigger*, *character frame* (Chłopicki 2017), *diffuse disjunction*, and *register humour* (Attardo 2001: 103-125) could not be accounted for by an analysis similar to that of examples (18), (19), and (20).

# 7 Visual humour

## 7.1 Graphic, audiovisual, and multimodal humour

Recent scholarship has suggested the direct expansion of the GTVH framework towards a unified theory of verbal and visual humour, or the linguistic and semiotic approaches to humour (Tsakona 2009). It seems uncontroversial that, "while linguistic theories in their formulation do not concern themselves with extra-linguistic matters, their underlying principles apply to all semiotic systems, and hence to multimodal and audiovisual texts" (Attardo 2020: 342). We hope our reformulation of script overlap in terms of constraints will make this unification more straightforward and evident. Thus, some of the incongruities in (18) (*King Lear*) and (19) (Monty Python) were not utterances, but visual stimuli. That there is no unsurmountable gap between verbal and non-verbal humour is made evident by the existence of multimodal humour. The following canned joke could be recast as a single-panel cartoon, with all but the two lines of dialogue replaced by visual stimuli.

(21) A group of bats is hanging from a branch with their heads down but one of them is standing up. A pair of bats next to him remark: "Sorry, what's up with him?" "I don't know, until two minutes ago he was fine then he fainted." (Contemporary joke)

In its cartoon version, the incongruity—the bat standing up—would be a visual stimulus in $F'_1$. Like (19), this is a classic instance of role-reversal. $f$ takes inputs from the set {*state of consciousness*, *body orientation in absolute terms*}, and outputs the value of the input property selection for a creature in a position contrary to normal. The input-output pair (*state of consciousness*, {*in a faint*, …}) belongs to $f$, but also to $f'$, where $f'$ takes input (*state of consciousness*, $_{bat}$), factoring in the particular species' worldview, which $f$ overlooks. For the second input, *body orientation in absolute terms*, the output of $f'$ is {*up*}, incongruous under $f$. This output would be implicated solely by visual stimuli in a cartoon rendition of (21).

As an example of audiovisual humour, obviously an important modality where the model should be tested, we present the following scene from the film *Kung Fu Panda*. Po, a bumbling, gluttonous kung-fu enthusiast, is admitted into a legendary martial-arts school by sheer luck. On his first day of training, he is given a thrashing and dismissed by his teacher and classmates as a freak with no potential. Po drifts sadly to the school's beautiful garden, where he is approached by Oogway, the stereotypical benign old master:

(22) A dejected Po stands under a peach tree in the moonlight.
Oogway approaches.
OOGWAY: I see you have found the Sacred Peach Tree of Heavenly Wisdom.
Po spins around, his face dripping with peach juice.
PO: *(mouth full)* Is that what this is? I am so sorry. I thought it was just a regular peach tree.
(*Kung Fu Panda*, 2008)

The gag depends entirely on music and visuals, specially the poetic image of Po under the Sacred Peach Tree in the moonlight, and his apparent demeanour seen from the back, suggesting that he is in a brooding mood. For input *placement*, $f$ outputs {*under the Sacred Peach Tree*, …}, i.e., the set of conventional places for a dejected hero to ruminate over his bad luck. But that input-output pair belongs to $f \cap^p f'$, where $f'$ takes input (*placement*, $_{Po's\ character\ traits}$), factors in Po's gluttony, which $f$ overlooks, and outputs the set of appropriate places for him to indulge in it, which includes *under a peach tree*. Thus, the visuals belong to $F_0 \cap F'_0$. When Po spins around and gives the audience the opportunity to discover the output of a second input, (*activity*,

*Po's character traits*), *f'* outputs {*eating fruit*, …}, incongruous under *f*, and the resulting verbal and visual stimuli, including his face dripping with peach juice, belong to $F'_1$ and not to $F_1$.

As any screenwriter knows, character-driven comedy is underpinned by a "hidden" predictive pattern that, even when less expected, never fails to factor in a known set of traits and idiosyncrasies defining the comic character, a form of the recurring mechanism characteristic of longer-format humour observed in (19) (Monty Python).

It should be clear by now that nothing prevents us from substituting the word "stimuli" for the word "utterances" in the description of the *F* sets in Figures 5 and 6, thus obtaining a mechanism of humour in any modality. Musical scales, dance steps, decorative patterns, and body language—to name but a few—are evidence that constraints are constantly being evoked by stimuli other than language. In fact, constraints are by definition needed if we as much as accept the occurrence of incongruity in non-verbal communication. When all stimuli in a communicative event are non-verbal, the interpretation of other auditory and visual stimuli completely displaces the interpretation of utterances, and we step out of multimodal and into purely visual and non-verbal humour, as in the constructed example with which we opened this article:

(2) A scene in a musical comedy à la *Singin' in the Rain* (1952). Gene Kelly, Donald O'Connor, and Debbie Reynolds have just moved into a new apartment. They paper the walls, hang new curtains, unpack their belongings, and so on, in the most varied and ingenious ways, singing and dancing all along. At one point one of them jumps on a table and holds a lightbulb way up high. The other two grab the table by the edge from opposite sides and turn it, until the lightbulb is screwed in in the ceiling.

For input *target object rotated*, *f* outputs {*lightbulb*}, the input-output pair (*target object rotated*, {*lightbulb*}) belonging to *f*∩*f'*, where *f'* takes input (*target object rotated*, $_{no\ efficiency}$), factoring in the degree of effectiveness sought, an argument which *f* overlooks to assume the customarily demanded efficiency. The visual stimuli implicating this output belong to $F_0 \cap F'_0$. When the input changes to (*rotating entity transmitting rotation*, $_{no\ efficiency}$), *f'* outputs {*whole body and surface it stands on*, …}, incongruous under *f*, and the visuals implicating this output belong to $F'_1$ and not to $F_1$.

## 7.2 Slapstick

Most importantly, above examples (2) and (21) (bats cartoon) illustrate the simultaneous evocation of *f* and *f'*, the relationship between them being logical, not temporal. Not only *f'* may be evoked before *f* is, as in (16) (Bassompierre joke) and (17) (the two atoms), but in (2) and (21) the two inputs that feed into both constraints may indeed be evoked simultaneously, indistinctly, or in any sequential order. Once this is contemplated, an account of slapstick comedy in terms of the mechanism of Figures 5 and 6 becomes unproblematic. The cream pie fight comedy routine, in which a familiar object—the pie—becomes something else—a weapon—seems like an illustrative case, for the overlapping of two design adaptations, one natural, the other imposed by the predicament the character is in, abounds in this kind of comedy. In a famous scene in *The Gold Rush* (1925), Charlie Chaplin, stranded in a remote cabin, tricks his hunger by boiling his shoe in a pot for a Thanksgiving dinner. As he feasts on his "meal" with gourmet-like flourishes, a succession of overlapping design adaptations can be seen at work, such as when he eats the shoelaces as if they were spaghetti. For input *physical properties*, *f* outputs the set of properties that make shoelaces adequate for tying one's shoes, {*thin*, *long*, *flexible*, *resistant*, …}. But this input-output pair belongs to $f \cap^p f'$ as well, where *f'* takes input (*physical properties*, $_{food}$) and outputs the set of properties that make shoelaces adequate for passing as what the overlooked argument specifies, {*thin*, *long*, *flexible*, *spaghetti-like*, …}. The stimuli implicating the spaghetti-like properties of shoelaces belong to the overlap $F_0 \cap F_0'$. But for input (*handling*, $_{food}$) the output of *f'* includes actions like *twirling them on fork*, as is done with real spaghetti, resulting in visual stimuli belonging to $F_1'$ and not to $F_1$. A similar analysis can be made for the shoe nails, which Chaplin delicately nibbles on as if they were the bones of a pheasant, and so on.

Bergson (1913 [1900]) explained humour as the perception of the mechanical in human life. Life is characterised by flexibility and agility, whereas mechanical movement is characterised by rigidity. A man falling is a comic situation because it shows the presence of a rigidity in life. We believe this duality, so typical of slapstick, can be explained as a particular case of the more general description afforded by the overlapping constraints model: the case in which the overt pattern predicts human-like movement, the covert pattern predicts mechanical movement, and both overlap implicitly in everyday human activities, for the human body, even when purposefully moved, is subject to the same physical laws inanimate objects are.

# 8 Falsifiability

The model we present can be said to be a good model of humour if it is predictive, i.e., if it assigns the property "humorous" to the same entities a human speaker would intuitively classify as "humorous". Our hypothesis should be tested against a larger corpus than the one we have used in this study and subjected to carefully designed psychological experiments. All that would be needed to falsify the theory is a communicative event perceived as humorous by human speakers, but which cannot be mapped onto the model, or one that can be mapped onto the model, yet it is not perceived as humorous by human speakers.

The second possibility seems more likely to us, but we believe that in all such cases we will be dealing with texts like:

(23) When the last tree is cut down, the last fish eaten, and the last stream poisoned, you will realize that you cannot eat money. (Internet-circulated Cree Indian saying).

(24) Being honest isn't a question of saying everything you mean. It's a question of meaning everything you say. (Quoted in Attardo and Raskin 1991: 305)

The seriousness of the context makes the psychological effect of both (23) and (24) more akin to having a revelation, or even an epiphany, than to getting a joke. Still, there is an element of playfulness to them that can justifiably be characterised as humour. Our mechanism, therefore, is most definitely *not* a model of jokehood (Ritchie 2018: 8), but a model of humour in its broadest sense. On closer inspection, though, witticisms not unlike (23) and (24) can properly be called "jokes" when uttered, by way of an example, by the fool-character in an Elizabethan play. It seems safe to say that jokes are a subset of the wider set of humorous utterances, but the characterisation of such subset may have to contemplate culturally shifting and historically evolving conventions and practices. It may well be that, like the class of "games", the class of "jokes" cannot be defined by finding a property shared by all jokes, but it comprises different sets of entities interrelated by "family resemblance" (Attardo 1994: 8-10; Wittgenstein 2009 [1953]: §66-71), in what would be an anti-essentialist theory of jokehood. This, however, falls outside of the scope of this study.

# 9 Conclusion

We have laid out a falsifiable mechanism of humour in any communicative modality. Our conviction that it describes the phenomenon of humour finds justification in the fact that it captures formally the intuition behind the words "incongruity" and "resolution": there cannot be an incongruity without a constraint being violated, and there cannot be a resolution without a second constraint modelling the situation "better" that the first. Nothing else is added, beyond the assumptions laid out in Section 3. Our decidedly abstract, all-encompassing definition of "constraint" as a correspondence between *any* two sets of data was needed to account for non-verbal humour, and instances of humour whose constituent elements are not linearly organised.

As a result, our schema may be characterised as a one-size-fits-all pair of gloves: it may not be the most direct way to describe the processing of a particular joke, but it is meant to map onto *any* humour-carrying communicative acts, and only these. Ultimately, our goal has been to answer the question, "what is humour".

Certainly, incongruity-resolution theories are not the only existing explanations of humour, but we believe current scholarship tends to agree that the other two historically important frameworks, release theories and superiority theories, do not answer the question, "what *is* humour", but the question, "what do people *use* humour for" (Attardo and Raskin 1991, 330-331). That is, they deal with communicative goals—disparagement—and psychological goals—release—of the use of humour in everyday life. As such, they deserve the serious consideration they still get from some scholars in the characterisation of *jokehood*, as opposed to that of *humour* in the broadest sense of the term.

By way of an example, superiority theories do not explain why certain presentations of a hostile thought are humorous, whereas others are not. Incongruity-resolution theories, on the contrary, can easily explain hostility, since the covert pattern may recover, for instance, a shared prejudice about another group not openly manifested in polite conversation, accounting for the frequent co-occurrence of humour and disparagement, and the collaborative engagement of a joke's audience to find the incongruent interpretation at the slightest opportunity afforded them (Veale 2004). Even when humour does not have a target of aggression, there is always a least one "victim", the hearer—or the hearer's own past self, before getting the joke—for the speaker proves to be intellectually skilful enough to entertain a pattern the hearer is unaware of. This explains the duelling character of many riddle and joke telling dynamics.

Pattern detection is what the human brain does best. The ability to construct good descriptions of the world and perceive relationships and analogies leads to better predictions, and in the evolutionary history of our species it must have led to higher survival and reproductive rates. Our description amounts to the basic idea that the process of pattern overlapping and substitution, even if not explicitly entertained by the hearer, is the essence of humour. From a biological point of view, it makes sense that games based on exercising and signalling our alertness and flexibility in pattern detection should have become a major source of pleasure in humans.